\documentclass[conference]{IEEEtran}
\usepackage{algorithm}
\usepackage{algpseudocode}
\usepackage{amsmath,amsthm,amsfonts}
\usepackage{graphicx}
\usepackage{nicefrac}

\usepackage{amsmath} 
\usepackage{booktabs} 
\usepackage[table]{xcolor} 
\usepackage{caption} 
\usepackage{float} 
\usepackage{multirow} 

\usepackage{glossaries}

\usepackage{cite}

\usepackage{array} 
\usepackage{adjustbox}
\ifCLASSINFOpdf
\else
\fi

\theoremstyle{plain}
\newtheorem{definition}{Definition}

\begin{document}
%
\title{Decentralized Federated Dataset Dictionary Learning for Multi-Source Domain Adaptation}

\author{\IEEEauthorblockN{Rebecca Clain \quad Eduardo Fernandes Montesuma \quad Fred Ngolè Mboula}
\IEEEauthorblockA{CEA-List, Université Paris-Saclay, F-91120 Palaiseau, France\\
rebecca.clain@cea.fr \quad eduardo.fernandesmontesuma@cea.fr \quad fred-maurice.ngole-mboula@cea.fr}}


%


\maketitle

\begin{abstract}
Decentralized Multi-Source Domain Adaptation (DMSDA) is a challenging task that aims to transfer knowledge from multiple related and heterogeneous source domains to an unlabeled target domain within a decentralized framework. Our work tackles DMSDA through a fully decentralized federated approach. In particular, we extend the Federated Dataset Dictionary Learning (FedDaDiL) framework by eliminating the necessity for a central server. FedDaDiL leverages Wasserstein barycenters to model the distributional shift across multiple clients, enabling effective adaptation while preserving data privacy. By decentralizing this framework, we enhance its robustness, scalability, and privacy, removing the risk of a single point of failure. We compare our method to its federated counterpart and other benchmark algorithms, showing that our approach effectively adapts source domains to an unlabeled target domain in a fully decentralized manner.
\end{abstract}


%
\IEEEpeerreviewmaketitle

\section{Introduction}
Machine learning models typically assume that training and test data come from the same distribution, which is rarely the case in real-world applications. This discrepancy, known as distributional shift, can significantly degrade model performance \cite{quinonero2008dataset}. In this context, Multi-Source Domain Adaptation (MSDA) has emerged as a strategy for adapting multiple heterogeneous labeled source datasets to an unlabeled target dataset \cite{montesuma2023learning}. Building on this concept, the decentralized MSDA addresses this challenge in a distributed setting, enabling models to leverage data from various sources (i.e., clients) without the need to aggregate the data centrally \cite{peng2019federated,liu2023co,feng2021kd3a}.  However, most decentralized MSDA approaches rely on a central server to coordinate the training process and update the model, introducing risks such as a single point of failure and potential bottlenecks. To mitigate these issues, removing the central server is a viable solution. Each client trains the model locally and directly shares updates with its peers. Our work builds upon this idea by implementing decentralized MSDA through a fully decentralized approach. Specifically, we extend the Federated Dataset Dictionary Learning (FedDaDiL) framework, presented in \cite{espinoza2024federated,montesuma2024}, by removing the necessity for a central server. FedDaDiL leverages Wasserstein barycenters to model the distributional shift across multiple clients, enabling effective adaptation while preserving data privacy. Fully Decentralizing this framework mitigates the limitations of traditional decentralized MSDA approaches, enhancing robustness, scalability, and privacy.
\par The rest of this paper is organized as follows: Section~\ref{Related_work} presents related works on decentralized federated learning and decentralized MSDA. Section~\ref{proposed_approach} introduces our fully decentralized approach of FedDaDiL for MSDA. Section~\ref{Experiments_and_discussion} presents our experiments on various domain adaptation benchmarks. Finally, Section~\ref{ccl} concludes this paper. 

 

\section{RELATED WORK}
\label{Related_work}
\subsection{Decentralized Federated Learning}

Decentralized federated learning is a collaborative approach where multiple clients train a shared model without a central server. Each client initializes its local model parameters, $\theta_i^{(0)}$, and performs local training to minimize its loss function, $\mathcal{L}_i(\theta)$. Clients exchange and aggregate model updates directly with peers. For client $i$, the aggregated model update from its neighbors $\mathcal{N}_i$ is computed as:
\[
\theta_i^{(t+1)} = \frac{1}{|\mathcal{N}_i|} \sum_{j \in \mathcal{N}_i} \theta_j^{(t)}.
\]

Several papers have proposed decentralized federated learning approaches \cite{kalra2021proxy,jiang2020bacomment,roy2019braintorrent,yuan2023decentralized,pakpahan2024peer}. However, these methods typically do not consider different domains, which limits their applicability for MSDA. In contrast, this paper introduces a novel decentralized federated approach specifically designed for MSDA, effectively managing distributional shifts across multiple source domains.

\subsection{Decentralized Multi-Source Domain Adaptation}
Federated learning frequently encompasses clients whose datasets exhibit disparate distributions, thereby violating the traditional assumption in machine learning that training and test data are i.i.d. Decentralized MSDA addresses a specific aspect of this challenge. It focuses on adapting models trained on multiple heterogeneous source domains to an unlabeled domain. Existing literature on decentralized MSDA includes the work of \cite{peng2019federated}, which proposes an approach that aligns representations learned across different nodes with the target node's data distribution through adversarial learning and feature disentanglement. Additionally, the authors of \cite{liu2023co} propose Federated Multi-Source Domain Adaptation on Black-box Models (B2FDA), which diverges from conventional approaches by avoiding the exchange of model parameters or gradients. Instead, it treats source nodes as 'black-box' models, sharing only soft outputs (the probabilities assigned by source models to each class) with the target node. Eventually, the author of  \cite{feng2021kd3a} perform decentralized MSDA through the knowledge distillation on models from different source domains. Despite their emphasis on privacy, these approaches still necessitate the aggregation of certain information on a central server, such as model parameters, gradient details or soft outputs, which introduces potential security vulnerabilities. In our paper, we propose to address decentralized MSDA without relying on a central server, thereby enhancing the robustness and privacy of the adaptation process.

\section{PROPOSED APPROACH}
\label{proposed_approach}
\subsection{Background}
Let's begin with an introduction to Wasserstein distance and Wasserstein barycenters. The Wasserstein distance, based on Optimal Transport (OT) theory, provides a robust framework for comparing distributions. Our approach leverages the Kantorovich formulation of OT~\cite{peyre2019computational,montesuma2023recent}.  Given empirical distributions $\hat{P}$ and $\hat{Q}$, represented by samples $x_i^{(P)} \sim P$ and $x_i^{(Q)} \sim Q$, we define:
\begin{equation*}
    \hat{Q}(x) = \frac{1}{n} \sum_{i=1}^{n} \delta(x - x_i^{(Q)})
\end{equation*}
and similarly for $\hat{P}(x)$. In this context $x_i^{(Q)}$ is call the support of $\hat{Q}(x)$. To compare distributions $P$ and $Q$ with $m$ and $n$ samples respectively, we consider the set of transport plans:
\begin{equation*}
    \Pi(P, Q) = \left\{ \pi \in \mathbb{R}_{+}^{n \times m} : \pi \mathbf{1}_m = \frac{1}{n} \mathbf{1}_n, \pi^T \mathbf{1}_n = \frac{1}{m} \mathbf{1}_m \right\}.
\end{equation*} 
Here, $\pi$ represents the plan for transporting mass from samples of $P$ to samples of $Q$. Given a cost matrix $C_{ij} = c(x_i^{(P)}, x_j^{(Q)})$, the Wasserstein distance is defined as
\begin{equation*}
    W_c(\hat{P}, \hat{Q}) = \min_{\pi \in \Pi(P, Q)} \sum_{i=1}^{n}\sum_{j=1}^{m}\pi_{ij}C_{ij},
\end{equation*}
where $c$ is a distance metric between samples. This distance facilitates the definition of a barycenter of distributions.

\begin{definition}
Given distributions \( P = \{P_k\}_{k=1}^{K} \) and weights \( \alpha \in \Delta^K \), the Wasserstein barycenter is defined as:
\begin{equation*}
    B^\star = B(\alpha; P) = \inf_{B} \sum_{k=1}^{K} \alpha_k W_c(P_k, B).
\end{equation*}
\end{definition}
The authors of \cite{montesuma2023learning} propose a novel framework for MSDA leveraging Wasserstein barycenters and Dictionary Learning (DiL). Traditionally, DiL decomposes a set of vectors $\{x_1, \ldots, x_N\}$ into a linear combination of atoms $\{p_1, \ldots, p_K\}$, weighted by representation vectors $\{\alpha_1, \ldots, \alpha_N\}$. The authors extend the DiL framework to handle empirical distributions. 
\newline 
\\
Given datasets $Q = \{\hat{Q}_{S_\ell}\}_{\ell=1}^{N_S} \cup \{\hat{Q}_T\}$, where $\{\hat{Q}_{S_\ell}\}_{\ell=1}^{N_S}$ represent the source domains and $\hat{Q}_T$ represents the target domain, their approach, called Dataset Dictionary Learning (DaDiL), learns atoms $P = \{\hat{P}_k\}_{k=1}^{K}$ and barycentric coordinates $A = \{\alpha_\ell \in \Delta^K\}_{\ell=1}^{N}$ such that:
\begin{equation*}
    (P^\star, A^\star) = \arg\min_{P, A} \frac{1}{N} \sum_{\ell=1}^{N} f_\ell(\alpha_\ell, P),
\end{equation*}
where $N = N_S + 1$, and $f_\ell$ is defined as
\begin{equation*}
    f_\ell(\alpha_\ell, P) = 
\begin{cases}
W_c(\hat{Q}_\ell, B(\alpha_\ell; P)) & \text{if } \hat{Q}_\ell \text{ is labeled}, \\
W_2(\hat{Q}_\ell, B(\alpha_\ell; P)) & \text{otherwise}.
\end{cases}
\end{equation*}
With $c$ a ground metric incorporating labels, and $W_2$ using the norm 2. This objective function is minimized with respect to the parameters $(X(P_k), Y(P_k))$ for each atom $P_k$ and the barycentric coordinates $\alpha_\ell$ for each domain. In short, DaDiL represents each dataset distribution $\hat{Q}_\ell$ as a labeled barycenter of the learned atoms $\mathcal{P} = \{\hat{P}_k\}_{k=1}^{K}$. 

In this context, two methods leverage the learned  dictionary, $(\mathcal{P}, \mathcal{A})$, to predict labels in the target domain: 
\begin{itemize}
    \item \emph{DaDiL-R} reconstructs labels from learned target barycenter.
    \item \emph{DaDiL-E} fits an Ensemble classifiers on the atoms distributions weighted by the target barycenters cooardinates.
\end{itemize}


Eventually, the authors of \cite{espinoza2024federated} propose FedDadiL, which extends DaDiL to a federated learning setting by distributing the optimization of $(\mathcal{P}, \mathcal{A})$ among multiple clients. A central server initializes atoms $P_g^{(0)}$ and send them to the clients. Each client then optimize their local versions of the atoms $P_\ell$ and barycentric coordinates $\alpha_\ell$ through $E$ local epochs while keeping $\alpha_\ell$ private. After each iteration, the server aggregates the optimized $P_\ell$ from all clients and redistributes the updated public $P_g$.

\subsection{Decentralized federated dataset dictionary learning}

\begin{figure}[ht]
\hspace{0.8cm}
\includegraphics[width=0.37\textwidth]{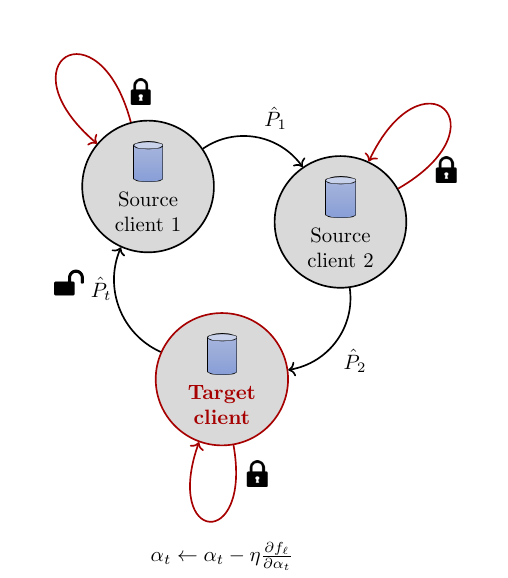}
        \caption{\small{De-FedDaDiL. Each client initializes atoms and, at each iteration, exchanges atom versions (\includegraphics[width=0.9em]{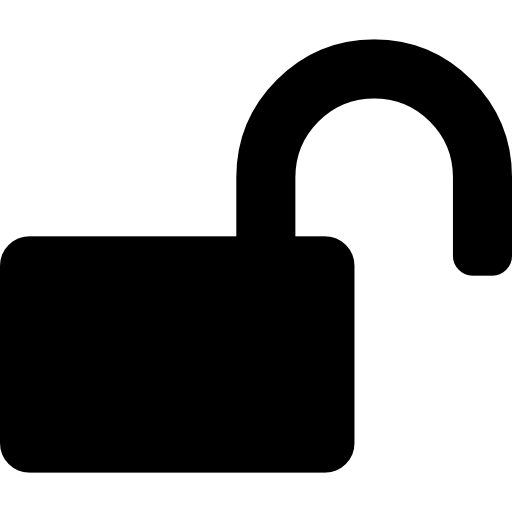}$\mathcal{P}$) with a randomly selected peer. Clients update their models based on their own and received version, while keeping barycentric coordinates private (\includegraphics[width=0.9em]{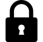}}$\alpha$). The learning process is collaborative and doesn't require a central server.}
    \label{fig:De_FedDaDil}
\end{figure}

In this paper, we propose a decentralized version of the FedDaDiL algorithm that removes the central server. In this decentralized FedDaDiL (De-FedDaDiL), the initialization and optimization of $(\mathcal{P}, \mathcal{A})$ are distributed across the clients, allowing for a fully decentralized operation (see~ Algorithm \ref{algo:Defed}). De-FedDaDiL is illustrated in Figure~\ref{fig:De_FedDaDil}.

In practice, each client initializes a local dictionary \{$P_{\ell}^{(0)}$, $\alpha_{\ell}^{(0)}$\}. At round $r$, every client (operating in parallel) shares its version of $P_{\ell}^{(r)}$ to a selected peer clients (\textproc{SelectPeer} in Algorithm~\ref{algo:Defed}) and receives a peer version from an other client, noted as $\mathcal{\tilde{P}}^{(r)}$. Each client aggregates the received $\mathcal{\tilde{P}}^{(r)}$ with their own version $P_{\ell}^{(r)}$ resulting in an intermediate version $P_{\ell}^{(r')}$ (\textproc{ClientAggregate} in Algorithm~\ref{algo:Defed}), which is then optimizes along with the barycentric coordinates $\alpha_{\ell}^{(r)}$ through $E$ local epochs (see Algorithm~\ref{algo:clientupdate}). As in the federated version the client's barycentric coordinates remain private. After a set number of iterations, the learning process stops. Each client has its own optimized version of the atoms and its associated barycentric coordinate ($P_{\ell}^{*}$, $\alpha_{\ell}^{*}$). It is important to note that, in contrast to FedDaDiL— where, by design, all clients share the same version of the atoms— each client in this approach maintains its own atom version. Our decentralized federated strategy, presented in Algorithm 1, is divided into three sub-routines: 

\noindent\textbf{\textproc{SelectPeer}}: At each iteration, every client (operating concurrently) selects a peer randomly and transmits its local version of the atoms $\mathcal{P}_{\ell}^{(r)}$. 

\noindent\textbf{\textproc{ClientAggregate}}: 
Following the execution of the \textproc{SelectPeer} function, each client possesses its own version of the atoms {$\mathcal{P}_{\ell}^{(r)}$, as well as the versions from one other peer $\mathcal{\tilde{P}}^{(r)}$}. It then aggregates these multiple local versions resulting in an intermediate version for this round, denoted as $\mathcal{P}_{\ell}^{(r')}$.
Following \cite{montesuma2024}, we aggregate these versions by averaging the supports of the atom distributions.

\noindent\textbf{\textproc{ClientsUpdate}}:
Based on the aggregation results $\mathcal{P}_{\ell}^{(r')}$, each client updates its parameters with respect to its own data ($\hat{Q}_{\ell}$). De-FedDaDiL follows the optimization procedure outlined in \cite[Alg. 2]{montesuma2023learning}. Each client optimizes $(\mathcal{P}_{\ell}^{(r')}, \alpha_{\ell})$ over $E$ steps, first splitting each $\hat{P}_{k}$ into \(B = \lceil \nicefrac{n}{n_{b}} \rceil\) batches of size \(n_{b}\). An epoch corresponds to an entire pass through the \(B\) mini-batches. The loss is calculated between mini-batches of \(\hat{P}_{k}\) and mini-batches of \(\hat{Q}_{\ell}\), as detailed in Algorithm~\ref{algo:clientupdate}. After each client step, \(\alpha_{\ell}\) is enforced to remain in \(\Delta_{K}\) by orthogonally projecting it onto the simplex.

\begin{algorithm}[ht]
\caption{De-fedDadiL}
\label{algo:Defed}
\begin{algorithmic}[1]
    \small
    \State \textbf{Initialization:} Each client $\ell \in \{1, \ldots, L\}$ initializes $\mathcal{P}_{\ell}^{(0)}$ and $\alpha_{\ell}^{(0)}$
    \For{each round $r = 1, \ldots, R$ \textbf{in parallel for each client} $\ell$}
        \State $client_{\text{selected}} = \textproc{SelectPeer}(L \setminus \{\ell\})$
        \State send $\mathcal{P}_{\ell}^{(r)}$ to $client_{\text{selected}}$
        \State receive $\mathcal{\tilde{P}}^{(r)}$ from some $\widetilde{client}$
        \State $\mathcal{P}^{(r')}_{\ell} \leftarrow \textproc{ClientAggregate}(\mathcal{P}^{(r)}_{\ell}, \mathcal{\tilde{P}}^{(r)})$
        \State $\mathcal{P}_{\ell}^{(r+1)}, \alpha_{\ell}^{(r+1)} \leftarrow \textproc{ClientUpdate}(\mathcal{P}^{(r')}_{\ell}, \alpha_{\ell}^{(r)})$
    \EndFor
\end{algorithmic}
\end{algorithm}

\begin{algorithm}[H]
\caption{ClientUpdate}
\label{algo:clientupdate}
\begin{algorithmic}[1]
    \State \textbf{Input:} Local atom $\mathcal{P}$. Set of weights $\alpha_{\ell} \in \Delta_{K}$. Number of local epochs $E$. Learning rate $\eta$.
    \For{local epoch $e = 1, \cdots, E$}
        \For{batch $b = 1, \cdots, B$}
            \State $f_{\ell}(\alpha_{\ell}; \mathcal{P}^{(e)}_{\ell}) = Wc(\hat{Q}_{\ell}, B(\alpha_{\ell}; \mathcal{P}^{(e)}_{\ell}))$
            \State $x^{(P_{k})}_{i} \leftarrow x^{(P_{k})}_{i} - \eta \frac{\partial f_{\ell}}{\partial x^{(P_{k})}_{i}} (\alpha_{\ell}, \mathcal{P})$
            \State $y^{(P_{k})}_{i} \leftarrow y^{(P_{k})}_{i} - \eta \frac{\partial f_{\ell}}{\partial y^{(P_{k})}_{i}} (\alpha_{\ell}, \mathcal{P})$
            \State $\alpha_{\ell} \leftarrow \text{proj}_{\Delta_{K}} (\alpha_{\ell} - \eta \frac{\partial f_{\ell}}{\partial \alpha_{\ell}} (\alpha_{\ell}, \mathcal{P}))$
        \EndFor
    \EndFor
    \State Client sets $(\alpha^{(r+1)}_{\ell}, \mathcal{P}^{r+1}_{\ell})\leftarrow (\alpha^{\star}_{\ell}, \mathcal{P}^{\star}_{\ell}) $
\end{algorithmic}
\end{algorithm}

In short, De-FedDaDiL functions in a fully decentralized manner by enabling the direct exchange and aggregation of local dictionary updates among peers. This approach supports the implementation of DaDiL \cite{montesuma2023learning} without a central server, ensuring robust and secure DMSDA.
\section{EXPERIMENTS AND DISCUSSION}

\label{Experiments_and_discussion}

\begin{table*}[h]
    \centering
    \caption{Experimental Results on decentralized MSDA benchmarks. $\star$ indicates results from~\cite{liu2023co}. $\uparrow$ denotes that higher is better.}
    
    \begin{minipage}{0.32\linewidth}
        \resizebox{1.\linewidth}{!}{%
            \begin{tabular}{lcccc>{\columncolor[gray]{0.9}}c}
                \toprule
                Algorithm & \scriptsize{Caltech} & \scriptsize{Bing} & \scriptsize{ImageNet} & \scriptsize{Pascal} & \scriptsize{Avg. $\uparrow$} \\
                \midrule
                FedAVG & 96.7 & 65.8 & 94.2 & 77.5 & 83.6\\
                FedProx & 96.7 & 65.8 & 93.3 & 76.7 & 83.1\\
                \midrule
                $f$-DANN & 96.7 & 64.2 & 87.5 & 80.0 & 82.1 \\
                $f$-WDGRL & 92.5 & 63.3 & 86.7 & 74.2 & 79.2 \\
                FADA & 95.0 & 64.2 & 90.0 & 74.2 & 80.9 \\
                KD3A & 93.3 & 69.2 & 95.5 & 73.3 & 82.8 \\
                Co-MDA & 94.2 & 65.0 & 91.5 & 78.0 & 82.2 \\
                \midrule
                FedDaDiL-E & 98.3 &  69.2 & 93.3 & 81.6 & 85.6\\
                FedDaDiL-R & 98.3 & 69.2 & 95.0 & 80.0 & 85.6\\
                \midrule
                De-FedDaDiL-E & 98.3 & 70.8 & 98.3 & 79.16 & \underline{86.6} \\
                De-FedDaDiL-R & 98.3 & 70.8 & 98.3 & 79.16 & \underline{86.6} \\
                \bottomrule
            \end{tabular}
        }
        \caption{ImageCLEF.}
    \end{minipage}\hfill
    \begin{minipage}{0.32\linewidth}
        \resizebox{0.9\linewidth}{!}{%
            \begin{tabular}{lccc>{\columncolor[gray]{0.9}}c}
                \toprule
                Algorithm & \scriptsize{Amazon} & \scriptsize{dSLR} & \scriptsize{Webcam} & \scriptsize{Avg. $\uparrow$} \\
                \midrule
                FedAVG & 67.5 & 95.0 & 96.8 & 86.4\\
                FedProx & 67.4 & 96.0 & 96.8 & 86.7\\
                \midrule
                $f$-DANN & 67.7 & 99.0 & 95.6 & 87.4 \\
                $f$-WDGRL & 64.8 & 99.0 & 94.9 & 86.2 \\
                FADA & 62.5 & 97.0 & 93.7 & 84.4 \\
                KD3A & 65.2 & 100.0 & 98.7 & 88.0 \\
                Co-MDA & 64.8 & 99.8 & 98.7 & 87.8\\
                \midrule
                FedDaDiL-E & 71.2 & 100.0 & 98.2 & 89.8 \\
                FedDaDiL-R & 70.6 & 100.0 & 99.4 & \underline{90.0} \\
                \midrule
                De-FedDaDiL-E & 68.3 & 99.7 & 98.7 & 88.9 \\
                De-FedDaDiL-R & 67.9 & 99.0 & 99.4 & 88.8 \\
                \bottomrule
            \end{tabular}
        }
        \caption{Office 31.}
    \end{minipage}\hfill
    \begin{minipage}{0.32\linewidth}
        \resizebox{1.05\linewidth}{!}{%
            \begin{tabular}{lcccc>{\columncolor[gray]{0.9}}c}
                \toprule
                Algorithm & \scriptsize{Art} & \scriptsize{Clipart} & \scriptsize{Product} & \scriptsize{Real-World} & \scriptsize{Avg. $\uparrow$} \\
                \midrule
                FedAVG & 72.9 & 62.2 & 83.7 & 85.0 & 76.0\\
                FedProx & 70.8 & 63.7 & 83.6 & 83.1 & 75.3\\
                \midrule
                $f$-DANN & 70.2 & 65.1 & 84.8 & 84.0 & 76.0 \\
                $f$-WDGRL & 68.2 & 64.1 & 81.3 & 82.5 & 74.0 \\
                FADA & - & - & - & - & - \\
                KD3A & 73.8 & 63.1 & 84.3 & 83.5 & 76.2 \\
                Co-MDA$^{\star}$ & 74.4 & 64.0 & 85.3 & 83.9 & 76.9\\
                \midrule
                FedDaDiL-E & 75.7 & 64.7 & 85.9 & 85.6 & \underline{78.0} \\
                FedDaDiL-R & 76.5 & 65.2 & 85.9 & 84.2 & \underline{78.0} \\
                \midrule
                De-FedDaDiL-E & 76.33 & 63.23 & 84.57 & 85.09 & 77.3 \\
                De-FedDaDiL-R & 76.13 & 63.45 & 84.12 & 84.86 & 77.1 \\
                \bottomrule
            \end{tabular}
        }
        \caption{Office-Home.}
    \end{minipage}

    \label{tab:fed_da_results}
\end{table*}

We compare De-FedDaDiL to other decentralized MSDA
strategies. An overview of the results is presented in Table~\ref{tab:fed_da_results}.
Specifically, we evaluate the methods on ImageCLEF~\cite{caputo2014imageclef}, Office 31~\cite{saenko2010adapting} and Office-Home~\cite{venkateswara2017deep}. We consider three state-of-the-art decentralized MSDA methods: FADA~\cite{peng2019federated}, KD3A~\cite{feng2021kd3a}, and Co-MDA~\cite{liu2023co}. Additionally, we compare De-FedDaDiL with its purely federated version FedDaDiL \cite{espinoza2024federated}. Furthermore, we evaluate adaptations of DA methods, such as $f$-DANN~\cite{ganin2016domain, peng2019federated} and $f$-WDGRL~\cite{shen2018wasserstein}.

The experimental results show that De-FedDaDiL is comparable to the federated version FedDaDiL and other reference algorithms. Across all datasets, De-FedDaDiL exhibits high accuracy, with De-FedDaDiL-E and De-FedDaDiL-R achieving average accuracies that are within 1-2\% of their federated counterparts.  This is significant, as it indicates that the fully decentralized method can achieve near-parity with federated approaches without the need for a central server. For instance, on Office 31 datasets, De-FedDaDiL variants show average accuracies that are very close to those of FedDaDiL, with differences of less than 1\% in most cases. Similarly, on the ImageCLEF dataset, De-FedDaDiL outperforms FedDaDiL and other benchmark algorithms. On the Office-Home, while De-FedDaDiL's performance is slightly lower than FedDaDiL, De-FedDaDiL still performs competitively against other benchmark algorithms. Eventually, in De-FedDaDiL, each iteration involves \( N \) communication exchanges, with each of the \( N \) clients interacting with one other client. In contrast, FedDaDiL requires \( 2 \times N \) exchanges per iteration, as each client must both send and receive updates from the central server. Given that both methods require a similar number of iterations, De-FedDaDiL is more efficient in terms of communication frequency. Additionally, by eliminating the central server as a single point of failure, De-FedDaDiL enhances system reliability.
\\ 
\newline 
In summary, the results demonstrate that De-FedDaDiL not only achieves performance levels comparable to the purely federated approach relying on a central server but also outperforms other state-of-the-art decentralized approaches, particularly in the ImageCLEF dataset. Additionally, De-FedDaDiL offers significant advantages such as enhanced system robustness and reduced communication overhead, making it a compelling and viable alternative for DMSDA tasks.
\subsection{Analyzing Consensus Among Clients' Atoms}
\begin{figure}[h!]
    \centering
    \includegraphics[width=\linewidth]{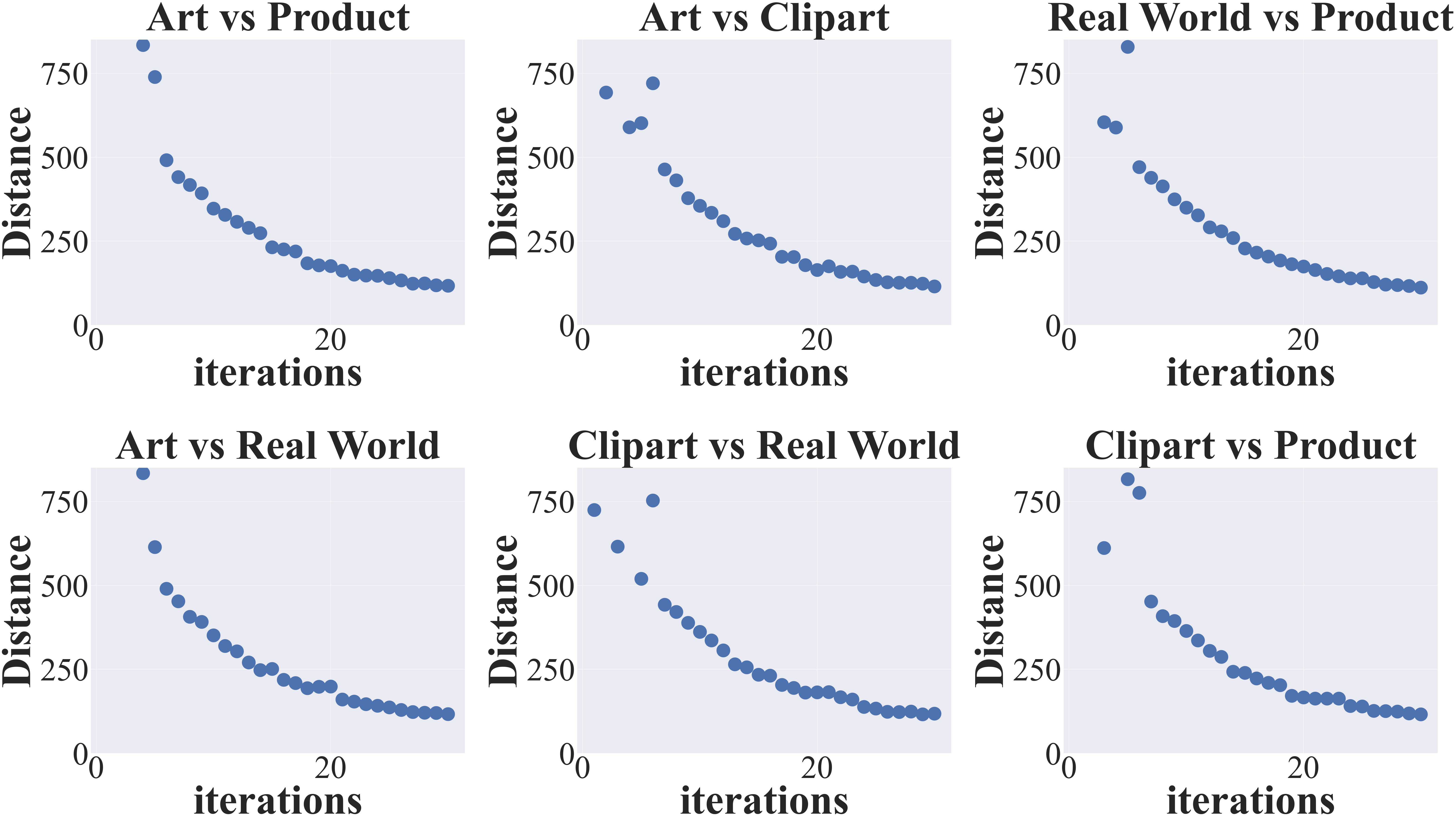}
    \caption{Max Wasserstein distance between the barycenters of the different domains over each training iteration - Target domain is art}
    \label{fig:consensus_analysis}
\end{figure}

In De-FedDaDiL, each client maintains its own optimized version of the atoms along with the corresponding barycentric coordinates. Although these atoms initially differ across clients, iterative sharing and averaging are expected to progressively align them, resulting in greater similarity over time. To assess consensus, we compute a hull of barycenters for each client’s atoms by generating multiple barycenters per iteration using various weight combinations. Specifically, we calculate the barycenters for each client by combining atoms with different weights and then compute the Wasserstein distance between barycenters that use the same weight combinations across clients. By tracking the maximum of these Wasserstein distances, we evaluate the degree of divergence and convergence among the clients' atoms over time.
 We illustrate our findings using results from the Office Home benchmark with Art as the target domain in Figure \ref{fig:consensus_analysis}.  The decreasing distance between barycenters over iterations indicates that clients are converging towards consensus.

\section{Conclusion}
\label{ccl}

This paper introduces De-FedDaDiL, an extension of the FedDaDiL approach to a fully decentralized setting. FedDaDiL uses Wasserstein barycenters and dictionary learning to efficiently adapt source domains to an unlabeled target domain. De-FedDaDiL allows this process to be performed in a fully decentralized manner, eliminating the need for a central server. Our results show that De-FedDaDiL achieves performance levels comparable to traditional federated methods, while improving system robustness, scalability, and privacy. By effectively addressing the challenges of decentralized multi-source domain adaptation, De-FedDaDiL proves to be both viable and efficient.


\section*{Acknowledgment}

This work has benefited from French State aid managed by the Agence Nationale de la Recherche (ANR) under France 2030 program with the reference ANR-23-PEIA-005 (REDEEM project)



%



\newpage
\bibliographystyle{IEEEtran}
\bibliography{acapps}

\end{document}